# Design and Experimental Study of Vacuum Suction Grabbing Technology to Grasp Fabric Piece

Ray Wai Man Kong[1], Mingyi Liu[2], Theodore Ho Tin Kong[3]
1. Adjunct Professor, Department of System Engineering, City University of Hong Kong, Hong Kong, Email: dr.raykong@gmail.com
SCM Director, Eagle Nice (International) Holding Ltd, Hong Kong
2. Department of Mechanical Engineering and Robotics, Guangdong Technion – Israel Institution of Technology, Shantou, China
3. Master of Science in aeronautical engineering, Hong Kong University of Science and Technology, Hong Kong

**Abstract** — The primary objective of this study was to design the grabbing technique used to determine the vacuum suction gripper and its design parameters for the pocket welting operation in apparel manufacturing. It presents the application of vacuum suction in grabbing technology, a technique that has revolutionized the handling and manipulation to grasp the various fabric materials in a range of garment industries. Vacuum suction, being non-intrusive and non-invasive, offers several advantages compared to traditional grabbing methods. It is particularly useful in scenarios where soft woven fabric and air-impermeable fabric items need to be handled with utmost care. The paper delves into the working principles of vacuum suction, its various components, and the underlying physics involved. Furthermore, it explores the various applications of vacuum suction in the garment industry into the automation exploration. The paper also highlights the challenges and limitations of vacuum suction technology and suggests potential areas for further research and development.

— — — — — — — — — ◆ — — — — — — — — — —

## 1 INTRODUCTION

Trought the intelligent manufacturing evolution, the handling in garment manufacturing to grasp the fabric into the machinery is involved in the manual load and unload operation to the machinery. This means that operators should work with machinery for garment sewing operations. The gripper and visual system are the focus aspects in applying automation technologies in instead of manual work.

It is the designer's responsibility to select the appropriate fabrics for their intended applications, based on the properties and characteristics of the fabrics. It is the responsibility of the fabric producer to provide as much information as possible to help the designers to make appropriate fabric selections. The information includes physical properties, style characteristics, tactile characteristics, utility characteristics, durability, and so on [2].

Utility characteristics in fabric are changes in the fit, comfort, and wearing functions of the garment when the fabric engages a mechanical thermal, electrical, or chemical force during the utilization of the garment. The two major types of utility characteristics are transmission and transformation. A transmission characteristic transmits mass or energy through the fabric. Transmission characteristics include:

- Air permeability (includes all gases and vapor)
- Heat transmission (thermal conductivity)
- Light permeability
- Moisture transmission

Manufacturing fabrics are related to fibres which are first spun together to make a yarn, which is then made into fabrics by being woven type and knit type.

Woven fabrics are created on a loom; warp threads are held under tension and the weft thread is woven between them, creating the patterns and design in the weave. There are different weaves available:

- plain - strong and hard wearing, eg calico and drill cotton, used for fashion and furnishing fabrics.
- twill - strong and drapes well with a diagonal pattern on the surface, used for jeans, curtains, and jackets.
- satin - fabric can be woven to give the surface a 'right' side with a shine, created by long floats on the warp or weft threads, and a 'wrong' side that is matt.
- pile - woven in two parts together that face each other and sliced apart down the centre once off the loom to create the pile. Some fabrics that are used for sporting garments contain elastomeric fibre.

Non-woven fabrics turn the fibres into fabrics without first spinning them but, instead, by felting or bonding them [3]. The fabric handling operations in the clothing industry can be divided in the following grabbing operations for: separation, grasping, picking, placing, and positioning. In the picking and placing process, the handling way which meet the expectations of complex movements comparable to those of the manual handled by hand, would be necessary, but currently, the existing systems are complex and generate little or no productivity improvement. On the other hand, the gripper for fabrics from holder to the automated machineries are required to grasp the soft fabric.

According to the universal gripper for fabrics – design, validation and integration from Yousef Ebraheem, Emilie Drean and Dominique Charles Adolphe of International Journal of Clothing Science and Technology in 2021, the vacuum technology offers a high reliability to handle air impermeable materials. The intrusion technology is reliable for the manipulation of high porosity materials, while the pinch technology shows good results for all soft fabrics when combined with the vacuum technology.

Compared to the various fabric, the knitted fabric consists of one yarn, woven fabric by multiple yarns in crisscross patterns. Woven fabric does not stretch across the width, but along the length. Knitted fabrics stretch easily across the width and slightly lengthwise. Woven fabrics structure is the most appropriate to apply the vacuum suction grabbing technology for automation.

The woven fabric can distinguish the major types between air permeable fabric type and non-permeable fabric type. The non-air permeable fabric is being focused on the vacuum suction grabbing technology and vacuum suction gripper.

In the Ref. [3], the automation machinery should correct the lifting of cut fabric from the adjoining fabric layer and take the fabric off. The picking must be firm but without any risk of damage to the fabric; separation must overcome the adhesive force between adjoining layers; taking-off requires lifting the separated fabric piece and holding separately. These actions must be followed by transfer of the cut fabric to a precisely defined position, with cross-positioning of the pieces to be joined and safe release of the cut piece without affecting the shape of the cut fabric pieces. Mechanical clamping grippers can lift the fabric piece, but there is not place the fabric piece in the required location at the high precision.

Followed the study of Universal gripper for fabrics – design, validation and integration from from Yousef Ebraheem as shown in the Ref. [3], the mechanical needle gripper has damaged on non-porous materials to make fabric tearing and deformation and the Risk to pick several layers at one time. It is difficult to apply the mechanical clamping gripper or needle gripper to take off the fabric piece and place it to the high precision location for automation.

This article is mainly focused on vacuum suction grabbing technology to apply a various number of vacuum suction grippers and various negative pressure supplies to the grippers.

## 2. LITERATURE REVIEW

Ref. [1] In the literature (Yuqi Liu, 2023), the mechanical soft gripper achieves a successful pickup in good condition, but a release fails to place fabric to the required area in the specific precision for the automated machine.

Ref. [3] In the literature (Yousef Ebraheem, 2021), the automatic handling of flexible materials has been the subject of much research over the last 20 years, with the most applications being in the textile industry. Several gripping devices have been developed to grasp a single layer of fabric from a pile of fabric and transfer it to a predefined location.

The pneumatic principle uses vacuum or differential pressure as operating principle and is therefore non-intrusive. Vacuum end effectors only require single surface access to the fabric piece for successful gripping. Most vacuum lifting principles use a compressed air powered vacuum gen-erator. This is because most fluid power or pneumatic manufacturers and distributors offer this technology instead of a motor-powered vacuum pump as they are small and cheap. The pneumatic grippers are specifically designed for the manipulation of textiles and fabric.

A good review of developing solutions to grab the fabric piece. This study makes the practical experiments and analysis to determine the vacuum suction gripper how to grasp the air impermeable fabric pieces up and placing the required area at the high precision for automation.

## 3. DESIGN ANALYSIS OF VACUUM SUCTION GRIPPER

The principle of vacuum suction is based on the creation and utilization of a pressure difference between two regions. This pressure difference is achieved by evacuating air from one region, creating a vacuum or partial vacuum, while the other region remains at ambient pressure or a higher pressure. The resulting pressure gradient generates a force that attracts and holds fabrics firmly against the suction surface.
In vacuum suction grabbing technology, a suction cup is used as the interface between the fabric, air impermeable fabric and the vacuum source. When the suction cup is pressed against the fabric, it excludes the air between the cup and the fabric surface. By evacuating the air through a vacuum pump or other means, a negative pressure (vacuum) is created within the suction cup. This negative pressure exceeds the atmospheric pressure outside the cup, creating a suction force that pulls the fabric securely against the suction cup.

The amount of suction force depends on the size of the suction cup, the difference in pressure between the inside and outside of the cup, and the size of holes of cup and the contract fabric area. The suction force must be sufficient to overcome any external forces acting on the fabric, such as gravity or friction, to enable effective fabric handling and manipulation.

Vacuum suction technology is widely used in various applications due to its simplicity, reliability, and adaptability to different types of air impermeable fabric surfaces. It is commonly found in automation systems, pick-and-place mechanisms, robotic grippers and other industrial equipment where precise and gentle handling of fabric cut pieces is crucial for the Apparel Manufacturing.

In grabbing technology, the formula of a vacuum suction cup usually involves several key parameters that determine the performance and efficiency of the suction cup. The following is a basic vacuum suction cup formula for calculating the vacuum pressure required to generate enough suction power to grip an object:

$$F = P * A \tag{1}$$

Thereinto:

* F is the suction force, and the unit is Newton (N).
* P is the vacuum pressure, and the unit is Pascal (Pa).
* A is the area of the suction cup, and the unit is square meters (m²).

This formula shows that the suction power (F) is the product of the vacuum pressure (P) and the area of the suction cup (A). Therefore, to increase the suction power, it can be achieved by increasing the vacuum pressure or increasing the suction cup area.

In practical applications, other factors need to be considered, such as the roughness of the surface of the object, air tightness, temperature, etc. These factors can affect the performance of the suction cups and therefore need to be carefully considered when designing a handling system.

In addition, in order to generate a vacuum, a vacuum pump is usually required. The choice of vacuum pump depends on factors such as the required vacuum, the flow rate, and the operating environment.

Overall, the formula for vacuum cups provides a basic framework for calculating the vacuum pressure required to generate enough suction power to grip an object. However, in practice, there are a variety of other factors that need to be considered to ensure the performance and reliability of the handling system.

Typical Vertical load
Load case– Suction cup horizontal, direction of force vertical

The fabric piece (in this case the fabric piece for the pocket facing 2.0g and pocket bag 2.5g is lifted from a plate as shown the weight of pocket facing and the weight of pocket bag. The fabric piece is lifted with an acceleration of 5 m/s² (no transverse movement).

$$F_{TH} = m \times (g + a) \times S \tag{2}$$
where, $F_{TH}$ = theoretical holding force [N], m = Weight [kg], g = Gravity [9.81 m/s²], a = Acceleration [m/s²], S = Safety factor

The vacuum generated by the vacuum generator can be adjusted by changing the compressed air pressure or flow rate, so it can supply the whole negative suction pressure required for different applications. In addition, the vacuum generator has the advantages of compact structure, easy installation, and maintenance, and can be widely used in various mechanical and automation equipment.

It should be noted that the vacuum generated by the vacuum generator is limited by the compressed air pressure and flow rate, so the maximum vacuum value cannot be unlimitedly high. Our engineer and I have selected the vacuum generator which selects the appropriate model according to the actual needs of the application and adjusts the compressed air pressure and flow rate according to the requirements to achieve the best vacuum effect.

For the vacuum grabbing technology, the air flow rate and pressure are required to calculate the vacuum grabbing force for grabbing the pocket bag, pocket facing, front and back of air impermeable fabric. The weight of the pocket bag, pocket facing, front and back are required to measure the weight of those fabric pieces. Based on the weight of those fabric pieces, the uplift grabbing theoretical holding force can be found by the vertical load in the formula [2].

In the vacuum suction grabbing force, the vacuum generator can supply the adequate negative air pressure to the suction gripper. If the negative air pressure to the suction gripper is greater than the required force, the vacuum gripper grasps more than one fabric piece of air non-permeable fabric piece. In another word, if the negative air pressure to the suction gripper is less than the required force, the vacuum gripper grasps nothing air non-permeable fabric piece.

The supply of negative air pressure value is required to be found out to grasp the fabric pieces. The Bernoulli equation can be applied to find out the supply vacuum suction pressure and force to satisfy the required loading force within the grabbing criteria.

Bernoulli equation is the conservation of energy principle appropriate for flowing fluids. The qualitative behavior that is usually labeled with the term "Bernoulli effect" is the lowering of fluid pressure in regions where the flow velocity is increased. This lowering of pressure in a constriction of a flow path may seem counterintuitive but seems less so when we consider pressure to be energy density. In the high velocity flow through the constriction, kinetic energy must increase at the expense of pressure energy.

While the Bernoulli equation is stated in terms of universally valid ideas like conservation of energy and the ideas of pressure, kinetic energy and potential energy, its application in the above form is limited to cases of steady flow. For flow through a tube, such flow can be visualized as laminar flow, which is still an idealization, but if the flow is to a good approximation laminar, then the kinetic energy of flow at any point of the compressed air can be modeled and calculated. The kinetic energy per unit volume term in the equation is the one which requires strict constraints for the Bernoulli equation to apply - it basically is the assumption that all the kinetic energy of the compressed air is contributing directly to the forward flow process of the compressed air. That should make it evident that the existence of turbulence or any chaotic compressed air motion would involve some kinetic energy which is not contributing to the advancement of the compressed air through the tube.

Applied the Bernoulli equation, the supply of negative pressure value can be found out to the appropriate negative pressure to vacuum suction gripper. It can grasp the woven fabric pieces successfully.

The supply of negative air pressure value is required to be found out to grasp the fabric pieces. The Bernoulli equation can be applied to find out the supply vacuum suction pressure and force to satisfy the required loading force within the grabbing criteria.

Bernoulli equation is the conservation of energy principle appropriate for flowing fluids. The qualitative behavior that is usually labeled with the term "Bernoulli effect" is the lowering of fluid pressure in regions where the flow velocity is increased. This lowering of pressure in a constriction of a flow path may seem counterintuitive but seems less so when we consider pressure to be energy density. In the high velocity flow through the constriction, kinetic energy must increase at the expense of pressure energy.

## 4. DESIGN OF EXPREIMENT FOR VACUUM SUCTION GRIPPER

Following the above theory, the design of experiment for vacuum suction gripper is designed to make it happen to make analysis of the number of grippers required and ap-propriate negative pressure for automation. Once the best solutions for grabbing the fabric pieces can be found by this study, the automated pick and place fabric piece can be applied to sewing machines, pocket welting machines, template sewing machines for garment manufacturing automation.

Suction cup vertical, direction of force vertical
Description of load case: The fabric piece (in this case the fabric sheet with the dimensions 2.5 x 1.25 m) is picked up from a holder and moved with a rotary motion at an acceleration of 5 m/s$^2$.

$$F_{TH} = (m/\mu) \times (g + a) \times S \qquad (3)$$

$F_{TH}$ = theoretical holding force [N], m = Weight [kg], g = Gravity [9.81 m/s$^2$], a = Acceleration [m/s$^2$] of the fabric to pick up and move it, μ = Friction coefficient, S = Safety

To generate vacuum suction for grabbing technology, a vacuum generator or vacuum pump is typically used. This equipment creates a negative pressure or vacuum by either extracting air from a sealed volume or reducing the pressure within a chamber. Here's a general outline of the steps involved in generating vacuum suction for grabbing technology:

1. Selection of Vacuum Pump or Generator:
Choose a vacuum pump or generator based on the required vacuum level, flow rate, and the size of the fabric to be grasped. There are various types of vacuum pumps available, including mechanical pumps, liquid ring pumps, and vacuum generators that use compressed air.

Name：Vacuum Generator
Band: Northmind
Model: CV-15HS
Parameter：
- Temp Range：0-60℃
- Pressure Range：1-6bar
- Nozzle Diameter：φ 1.5mm
- Setup Pressure：5bar
- Flow Rate：63L/min
- Negative Pressure：-92 kPa
- Air Consumption：100L/min

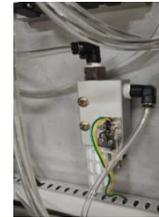

**Fig. 1.** The photo of Vacuum Generator

2. Design of the Vacuum Cup:
The vacuum cup, also known as a suction cup, is the component that creates the seal and attaches to the fabric. It is typically made of a flexible material that can conform to the shape of the fabric. The design of the vacuum cup should consider the size, shape, and surface characteristics of the fabric to be grasped.

Name：Vacuum Suction Cup
Band: HOFUJNG
Model:M8-L90
Parameter:
- Length：90mm
- Moving Distance：25mm
- Screw Lock: M8
- Cup diameter: 2mm

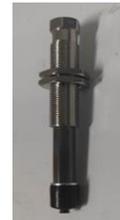

**Fig. 2.** The photo of vacuum suction cup

3. Connecting the Vacuum Pump to the Vacuum Cup:
Use a vacuum hose or tubing to connect the vacuum pump or generator to the vacuum cup. Ensure that the hose is long enough to reach the fabric and that it can withstand the vacuum pressure.

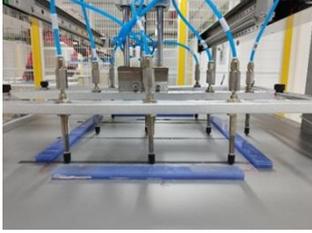

**Fig. 3.** The photo of Vacuum and nozzle clamp on the gripper frame.

## 5. MECHANICAL AND FLUID ANALYSIS FOR VACUUM SUCTION GRIPPER

<u>Load case– Suction cup vertical direction of force vertical</u>
The fabric piece (in this case the fabric piece for the pocket facing 2.0g and pocket bag 2.5g is lifted from a plate. The fabric pieces are lifted with an acceleration of 5 m/s² (no transverse movement).

FTH = (m/μ) × (g + a) × S        (4)
FTH = theoretical holding force [N]
m = Weight [kg]
g = Gravity [9.81 m/s2]
a = Acceleration [m/s2] of the fabric to pick up and move it
μ = Friction coefficient
S = Safety

The vertical loading force of pocket bag:
FTH = (2.5 x 10-3 kg/0.5) x (9.81 m/s² + 5 m/s2) x 2
FTH = 0.148 N

The vertical loading force of pocket facing:
FTH = (2.0 x 10-3 kg/0.5) x (9.81 m/s² + 5 m/s²) x 2
FTH = 0.118 N

The vertical loading force of the pocket bag and pocket facing are 0.148 N and 0118 N accordingly based on the fabric weight of the pocket bag and pocket facing accordingly.

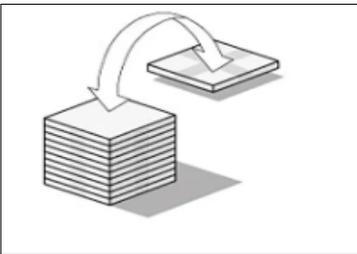

**Dia. 1.** Schematic diagram of vertical load for the pick and place motion

From the equation [1], the suction pressure is required to grasp up the pocket facing and pocket bag accordingly.
F = P * A
P = F / A
Referring to the suction cup, the diameter is 2mm, the required uplift pressure is calculated as shown below equation:
D = 2mm
A = (D/2) ²π
A = (2 x 10-3 m /2) ²π
A = (10-6 m) π
A = 3.1415 x 10-6 m

F = 0.148 N (for pocket bag)
F = 0.118 N (for pocket facing)

P = F/A
P = 0.148 N / 3.1415 x 10-6 m
P = 47,111Pa (for pocket bag)

P = F/A
P = 0.118 N / 3.1415 x 10-6 m
P = 37,561Pa (for pocket facing)

There into:
* F is the suction force, and the unit is Newton (N).
* P is the vacuum pressure, and the unit is Pascal (Pa).
* A is the area of the suction cup, and the unit is square meters (m²).

The required uplift pressure is 47.1kPa for the pocket bag and 37.5kPa pocket facing accordingly.

Bernoulli's equation relates the pressure, speed, and height of any two points (1 and 2) in a steady streamline flowing air density. Bernoulli's equation is usually written as follows:
$$P_1 + \frac{1}{2}\rho v_1^2 + \rho g h_1 = P_2 + \frac{1}{2}\rho v_2^2 + \rho g h_2 \quad (5)$$

For the automated machinery design, the equipment is reformatted to the below following:
$$\frac{P_1}{\rho g} + \frac{1}{2g}v_1^2 + h_1 + H_{pump} = \frac{P_2}{\rho g} + \frac{1}{2g}v_2^2 + h_2 + H_{Loss} + H_{Turbine} \quad (6)$$

Whereas the pump head, $H_{pump}$ is zero because there is the compressed air which initiates negative pressure from the vacuum pressure generator in $P_1$. The grabbing pressure in the vacuum suction gripper is $P_2$.
$$P_1 - P_2 = \rho \frac{(v_2^2 - v_1^2)}{2} + \rho g[(h_2 - h_1) - H_{pump} + H_{Loss} + H_{Turbine}]$$
$$(7)$$

At 101.325 kPa (abs) and 20 °C (68 °F), air has a density (ρ) of approximately 1.204 kg/m³ according to the International Standard Atmosphere (ISA). The gravity value, g on Earth is 9.8m/s². Kinematic viscosity of air at 20°C is given to be 1.6 × 10-5 m²/s.

The following formula is used to calculate the total flow rate in parallel pipes.

$$Q_{generator} = Q_{B1} + Q_{B2} + \cdots = \sum_1^n Q_n \qquad (8)$$

Where $Q_{B1}$ is the flow rate of branch 1 of pipe, $Q_{B2}$ is the flow rate of branch 2 of pipe … etc.

In order to determine the Flow Rate represented as Q, we must define both the volume V and the point in time it is flowing past represented by t, or Q = V/t. Additionally Flow rate and velocity are related by the equation Q = Av where A is the cross-sectional area of flow and v is its average velocity.

In a series of air pipes, by Bernoulli equation, there is the same pipe in the horizontal level to reduce the pipe diameter. The $h_1$ is the same height level of $h_2$. $A_1$ is calculated from the initial section area by the inner 5.2mm diameter of the air pipe. $A_2$ is calculated as the section area in the last air pipe as inner 2.0mm diameter.

$A_1$ = (5.2mm/2)² x π = 2.123 x 10⁻⁵ m²

$A_2$ = (2.0mm/2)² x π = 3.141 x 10⁻⁶ m²

Referring to Bernoulli's equation (5), in the series of air pipes, we can find out the velocity to relate to the pressure variation.

$$P_1 + \frac{1}{2}\rho v_1^2 + \rho g h_1 = P_2 + \frac{1}{2}\rho v_2^2 + \rho g h_2$$

Reformat the equation (5), there is shown below equation.

$$P_1 - P_2 = \frac{1}{2}\rho(v_2^2 - v_1^2) \qquad (9)$$

In the continuity equation, the conservation of mass is shown below.

$$\rho A_1 v_1 = \rho A_2 v_2 \qquad (10)$$

Applying the equation (10) into the equation (9), we have reformatted the equation.

$$P_1 - P_2 = \frac{1}{2}\rho v_1^2 \left(\frac{A_1^2}{A_2^2} - 1\right) \qquad (11)$$

$$P_1 - P_2 = \frac{1}{2} 1.204 \text{ kg/m}^3 (37.14 m/sec)^2 \left(\frac{(2.123 \times 10^{-5})^2}{(3.141 \times 10^{-6})^2} - 1\right)$$

$$P_1 - P_2 = \frac{1}{2} 1.204 \text{ kg/m}^3 (37.14 m/sec)^2 \left(\frac{(2.827 \times 10^{-5})^2}{(3.142 \times 10^{-6})^2} - 1\right)$$

$P_1 - P_2 = 830.3 \ (44.6) = 37,018.$ Pa

$\Delta P_{Loss} = P_1 - P_2 = 37.0 kPa$

There will be a pressure drop when pipes are connected in parallel. The pressure drop occurs because the air flowing through each parallel branch encounters resistance, which is caused by factors such as friction within the pipe walls, bends in the pipe, and changes in pipe diameter.

The pressure loss is 37kPa from the 5.2mm inner diameter of the air pipe to the 2mm diameter of air pipe. The maximum supply of negative is dropped from -92kPa to -55kPa.

The demamd of grabbing pressure for the pocket bag and pocket facing in woven fabric (100% Polyester; Plain Waves; Textile-woven) is 47.1kPa and 37.5kPa accordingly. The net supply vacuum pressure to gripper is -55kPa which can supply the sufficient negative pressure to grasp the pocket facing and packet bag for automated sewing mahine.

## 6. ANALYSIS OF GRABBING POSITION FOR FABRICS

**Ref. 5** Goran Cubric et al in 2012 have shown that the transfer of fabric required inlet pressure on the vacuum gripper a minimum of 5 bar. It has also been found that the application of this vacuum gripper is not suitable for taking one layer of fabric from a material bundle.

Transfer fabric piece with vacuum, such as taking out of a bundle and transfer to another location, there are very practical problems. Namely, the separation of individual fabric pieces for various reasons is not achieved easily.

Part of the fabric bundle held together by static electricity, partly due to the vacuum created between them, and in textile materials due to its permeability, vacuum engages the fabric piece below the uppermost of which catches.

The practical test was performed so that on the flat surfaces of the table base laid five to ten layers of woven fabric pieces and the vacuum gripper is set in the 2cm distance of fabric edge. (Dia. 2). Inlet negative pressure pressure to catch is changing from $P_1$ to $P_2$. The $P_1$ is the minimum grabbing negative pressure to grasp one piece of fabric pieces from the bundle. $P_2$ is the maximum grabbing negative pressure to grasp no more than one piece of fabric pieces from the bundle.

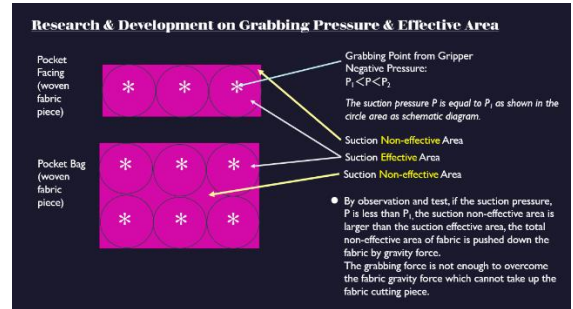

**Dia. 2.** Schematic diagram of grabbing force and position

The experiment is designed to keep the same distance 2cm between gripper point and edge of fabric piece based on the pocket facing is a rectangular shape in 4cm length and 12cm width. The pocket bag is a rectangular shape in 8cm length and 12cm width. The grabbing point is located in the center of a circle in the 2cm radius of the pocket facing and in the same 2cm ra-dius circle of the pocket bag.

Because the greater air suction pressure can pass through the structure of woven fabric, it can suck more than one layer of fabric pieces. For automation, the requirement of grabbing fabric pieces for an automated welting machine is only allowed to grasp one layout of fabric piece to pick up over 20cm and move horizontally 50cm to the required location of welting machine.

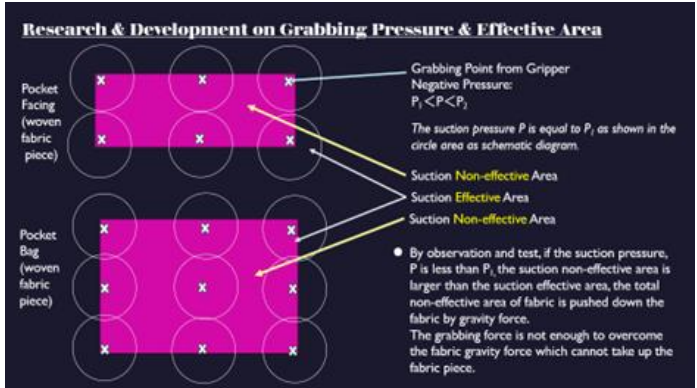

**Dia. 3.** Schematic diagram of grabbing force and position of grabbing point on the fabric if grabbing pressure is greater than min. pressure and less than maximum grabbing pressure

Referring to the Goran Cubric et al experiment, the grabbing position either in the corner of fabric or center of fabric affected the grabbing result (pass or fail). The grabbing point is an essential consideration to affect the result.

This study is exploring the new concept of grabbing pressure, and its related area based on the value of suction pressure as suction effective area as circle shape of schematic diagram in the Dia. 2 and Dia. 3. It is called a Vacuum Grabbing Pressure Theoretical Circle (VGTC).

If the grabbing point is pointing to the corner or edge of the fabric piece, the center of the suction effective area should be shifted to the corners or edge of the fabric piece. After shifting the grabbing point, the suction effective area is shifted to covered to outside of fabric; therefore, the non effective suction area exists on the fabric piece. The ratio of non-effective suction area and effective suction area is en-larged to make it fail during grabbing as shown in the Dia. 4.

Therefore, after shifting the grabbing point from the center of fabric piece (1cm distance to the edge or corners) to the edge and corner of fabric piece, the minimum grabbing negative pressure, $P_1$ increases to the higher value depend on the shape of fabric piece and grabbing point distance on the edge of fabric piece or whether keep the same distance of fabric piece.

The grabbing points, grabbing suction pressure, number of grippers, shape of fabric, fabric weight and fabric material are the factors that affect whether success to pick up the fabric piece.

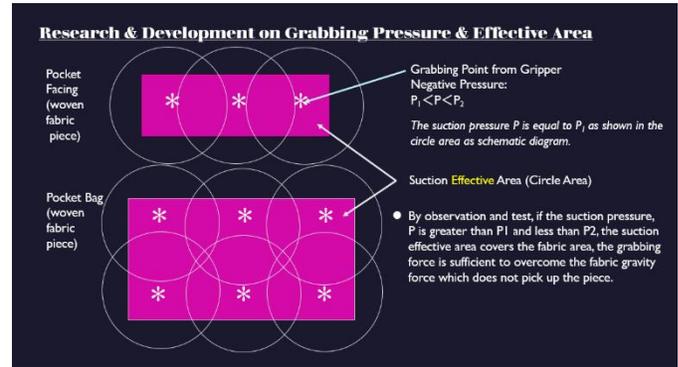

**Dia. 4.** Schematic diagram of grabbing force and position of grabbing point on the edge and corner of fabric piece if grabbing pressure is greater than min. pressure and near maximum grabbing pressure

## 7. TEST RESULT

Table 1 shows the results of capturing one layer of fabric piece lifting and holding up to the required location. Observing the results of capturing one layer of woven fabric with vacuum gripper within the range of grabbing pressure, P and the number of gripper to grasp the edge of pocket facing and pocket bag as shown Dia. 5.

TABLE 1
VACUUM GRABBING TEST FOR WOVEN FABRIC PIECES

| Test Lot /项目 | Fabric Piece Application 裁片使用 | Fabric Code 物料号 | Fabric 布种 | No. of Gripper 需求数量 | Outline rectangular Length & Width (cm) | Supply Negative Air Pressure 供抽吸气压 | Result (pass / fail) 结果（通过/失败） |
|---|---|---|---|---|---|---|---|
| 1 | Pocket Bag | 713993 | 100%Polyester; Plain Weave; TEXTILE-WOVEN | 6 | 26cm x 19cm | -55kPa | 通过 Pass |
| 2 | Pocket Bag | 1014845 | 68% Polyester, 32% Nylon; Taffeta; Plain Weave;TEXTILE-WOVEN; TEXTILE-WOVEN | 12 | 30cm x 36cm | -55kPa | 通过 Pass |
| 3 | Pocket Bag | 1004608(TW) | 96% Nylon (Mechanically Recycled), 4% Elastane; TEXTILE-WOVEN | 6 | 26cm x 19cm | -55kPa | 通过 Pass |
| 4 | Pocket Bag | 701696(CN) | 100% Polyeste; TEXTILE-WOVEN; Satin/Sateen | 6 | 26cm x 19cm | -55kPa | 通过 Pass |
| 5 | Pocket Bag | 602151(TW) | 100% Polyester (Recycled); Taffeta; TEXTILE-WOVEN | 12 | 30cm x 36cm | -55kPa | 通过 Pass |
| 6 | Pocket Bag | 000110 (CN) | 100% Nylon; Taffeta; TEXTILE-WOVEN | 8 | 26cm x 19cm | -55kPa | 通过 Pass |
| 7 | Pocket Facing | 713993 | 100%Polyester; Plain Weave; TEXTILE-WOVEN | 6 | 26cm x 5cm | -55kPa | 通过 Pass |
| 8 | Pocket Facing | 1014845 | 68% Polyester, 32% Nylon; Taffeta; Plain Weave;TEXTILE-WOVEN; TEXTILE-WOVEN | 6 | 30cm x 5cm | -55kPa | 通过 Pass |
| 9 | Pocket Facing | 1004608(TW) | 96% Nylon (Mechanically Recycled), 4% Elastane; TEXTILE-WOVEN | 6 | 26cm x 5cm | -55kPa | 通过 Pass |
| 10 | Pocket Facing | 701696(CN) | 100% Polyeste; TEXTILE-WOVEN; Satin/Sateen | 6 | 26cm x 5cm | -55kPa | 通过 Pass |
| 11 | Pocket Facing | 602151(TW) | 100% Polyester (Recycled); Taffeta; TEXTILE-WOVEN | 6 | 30cm x 5cm | -55kPa | 通过 Pass |
| 12 | Pocket Facing | 000110 (CN) | 100% Nylon; Taffeta; TEXTILE-WOVEN | 8 | 26cm x 5cm | -55kPa | 通过 Pass |

**Table 1** Vacuum grabbing test for woven fabric pieces.

Observing the results of capturing one layer of woven fabric piece with vacuum gripper (Table. 1) show that different sharp, size and material of fabric pieces test the appropriate number of grippers to get the pass result based on the Vacu-um Grabbing Pressure Theoretical Circle.

The result of the vacuum grabbing test can prove a measurement of the radius of the single grabbing circle and appropriate grabbing pressure to pick up the small portion of fabric piece as shown in Dia. 5. Using the small portion of fabric piece, the result of radius of VGTC can be found out under the appropriate range of grabbing pressure.

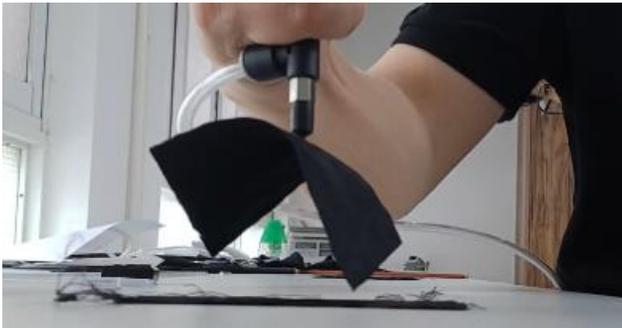

**Dia. 5.** Photo of single grabbing test to determine the grabbing pressure and radius of vacuum grabbing pressure theoretical circle.

The radius figure is the distance between grippers to locate on the grabbing frame for grabbing the required shape of fabric piece. The grabbing frame can keep the vacuum grippers to meet up the required location to pick the fabric piece in the new design of automated welting machine as shown the Dia. 6.

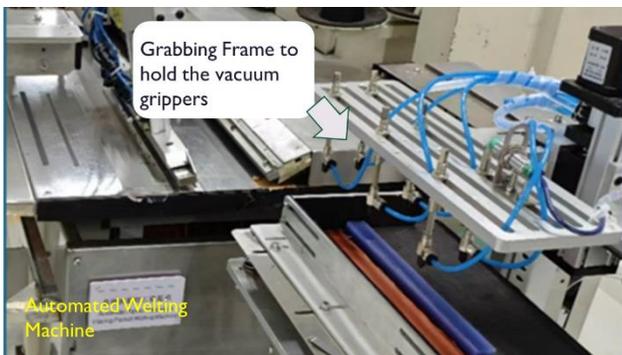

**Dia. 6.** Photo of grabbing frame to hold the vacuum grippers for automated welting machine

This article uses an experiment to test and verify theories in grabbing technology as well as Vacuum Grabbing Pressure Theoretical Circle. These experiments can range from simple observations or measurements to more complex setups involving specialized equipment or technologies.

The vacuum grabbing experiment that is often used to test the grabbing pressure is the controlled experiment. In a con-trolled experiment, the engineer manipulates one or more variables in grabbing pressure and measures the effect on a dependent variable to the weight, shape and material of wo-ven fabric piece. This allows them to determine the cause-and-effect relationship between the variables and to test whether a Vacuum Grabbing Pressure Theoretical Circle theory accurately predicts the results of the experiment.

The experiments that are commonly used to calculate the number of grippers and distribution of grippers to cope of fab-ric piece requirements which include simulations, observa-tions of natural phenomena, observation and measurements.

Regardless of the experiment being conducted, the goal is to gather data that can be used to support the workable solution with Vacuum Grabbing Pressure Theoretical Circle theory as explained by the observations.

## 8. CONCLUSION

A conclusion can validate the supply of vacuum negative pressure from the vacuum generator to cope with the demand of grabbing pressure to gripper. The grabbing force of the vacuum suction gripper can pick up the fabric piece. The Vacuum Grabbing Pressure Theoretical Circle and its simple test can determine the appropriate range of vacuum suction pressure to pick up one layer of fabric piece from the fabric bundle.

Furthermore, the radius of Vacuum Grabbing Pressure Theoretical Circle can apply to the distance between vacuum suction gripper to set up in the grabbing frame. Based on the shape of the fabric piece, the appropriate number of grippers can be calculated to set up on the grabbing holder which keeps the distance between grippers according to the radius of VGPTC. There is success to grasp the fabric piece from one location to another location as provided by the solution of the robotic pick and place solution for garment.

The breakthought the problem how to apply the vacuum suction gripper to grasp the soft fabric can overcome problems to apply the methodology and steps to grasp the fabric piece to automated machine by robotic gripper for the garment manufacturing automation.

## ACKNOWLEDGMENT

Prof. Kong Wai Man, Ray wishes to thank Professor CY Dang and Professor Cheng Liu from City University of Hong Kong, our assistance engineers from Eagle Nice (International) Holding Ltd.

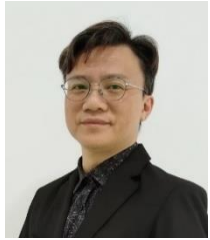

**Ray Wai Man Kong** (Senior Member, IEEE) Hong Kong, China. He received the Bachelor General Study degree in the Open University of Hong Kong, Hong Kong in 1995. He received the MSc degree in Automation Systems and Engineering and the Engineering Doctorate degree from City University of Hong Kong, Hong Kong in 1998 and 2008 respectively.

From 2005 to 2013, he was the OPERATION DIRECTOR with Automated Manufacturing Limited, Hong Kong. From 2020 to 2021, he was the CHIEF OPERATING OFFICIER (COO) with Wah Ming Optical Manufactory Ltd, Hong Kong. He is currently a SC DIRECTOR with Eagle Nice (International) Holdings Limited, Hong Kong. He holds an appointment, Adjunct Professor of System Engineering Department with the City University of Hong Kong, Hong Kong. He has published an Incremental Model-based Test Suite Reduction with Formal Concept Analysis in Journal of Information Processing Systems June 2010. ISSN: 2092-805X in Korea Information Processing Society. His research interest focuses on the Intelligent Manufacturing, Automation, Maglev Technology, Robotic, Mechanical Engineering, Electronics and System Engineering for industrial factory.

Prof. Dr. Kong Wai Man, Ray is Vice President of CityU Engineering Doctorate Society, Hong Kong and Consulting Member of Doctors Think Tank Academy, Hong Kong. He has published 5 intellectual properties and patent in China.

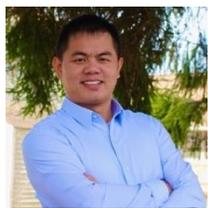

**Mingyi Liu** (Member, IEEE) received the M.S., from School of Instrumentation and Optoelectronic Engineering, Beihang University, Beijing, China in 2014 and Ph.D. Department of Mechanical Engineering, Virginia Tech, Blacksburg, U.S.A. in 2020. From May 2020 to March 2023, he was with the Center for Energy Harvesting Materials and Systems, Department of Mechanical Engineering, Virginia Tech, Blacksburg, as a Postdoctoral Fellow. He is currently assistant professor of Mechanical Engineering and Robotics, Guangdong Technion – Israel Institution of Technology (GTIIT), Shantou, Guangdong, China.

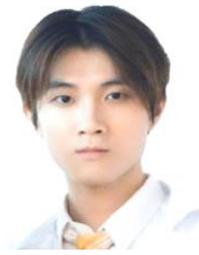

**Theodore Ho Tin Kong** (MIEAust, Engineers Australia) received his Bachelor of Engineering (Honours) in mechanical and aerospace engineering from The University of Adelaide, Australia, in 2018. He then earned a Master of Science in aeronautical engineering (mechanical) from HKUST - Hong Kong University of Science and Technology, Hong Kong, in 2019.

He began his career as a Thermal (Mechanical) Engineer at ASM Pacific Technology Limited in Hong Kong, where he worked from 2019 to 2022. Currently, he is a Thermal-Acoustic (Mechanical) Design Engineer at Intel Corporation in Toronto, Canada. His research interests include mechanical design, thermal management and heat transfer, and acoustic and flow performance optimisation. He is proficient in FEA, CFD, thermal simulation, and analysis, and has experience in designing machines from module to heavy mechanical level design.